\documentclass[a4paper,fleqn]{cas-dc}

\usepackage[authoryear,longnamesfirst]{natbib}
\usepackage{graphicx}
\usepackage{latexsym}
\usepackage{subcaption}
\usepackage{amsmath}
\usepackage{pifont}
\usepackage{multirow}
\usepackage{amsfonts}
\usepackage{makecell}
\usepackage{diagbox}
\usepackage{booktabs}
\usepackage{amssymb}
\usepackage{threeparttable}

\def\tsc#1{\csdef{#1}{\textsc{\lowercase{#1}}\xspace}}
\tsc{WGM}
\tsc{QE}
\tsc{EP}
\tsc{PMS}
\tsc{BEC}
\tsc{DE}

\begin{document}
\let\WriteBookmarks\relax
\def\floatpagepagefraction{1}
\def\textpagefraction{.001}

\shorttitle{Learning Robust Named Entity Recognizers From Noisy Data With Retrieval Augmentation}

\shortauthors{C. Ai et~al.}

\title [mode = title]{Learning Robust Named Entity Recognizers From Noisy Data With Retrieval Augmentation}
\tnotemark[1]

\tnotetext[1]{This document is the results of the research
   project funded by the National Natural Science Foundation of China (61976139) and Alibaba Group through Alibaba Innovative Research Program.}

%
\author[1]{Chaoyi Ai}



\ead{aichaoyi1998@gmail.com}

\credit{Methodology, Software, Data curation, Writing – original draft}

\affiliation[1]{organization={School of Information Science and Technology, ShanghaiTech University},
    addressline={No. 393 Huaxia Middle Road},
    city={Shanghai},
    postcode={201210},
    country={China}}

\author[2]{Yong Jiang}
\cormark[2]
\ead{yongjiang.jy@alibaba-inc.com}
\ead[URL]{http://jiangyong.site/}

\credit{Conceptualization, Methodology, Funding acquisition}

\author[2]{Shen Huang}
\ead{pangda@alibaba-inc.com}
\credit{Conceptualization, Methodology}

\author[2]{Pengjun Xie}
\ead{chengchen.xpj@alibaba-inc.com}
\credit{Supervision, Funding acquisition}

\affiliation[2]{organization={Alibaba Group},
    addressline={No. 969, Wen Yi West Road, Yuhang District},
    city={Hangzhou},
    postcode={311121},
    country={China}}

\author[1]{Kewei Tu}
\cormark[1]
\ead{tukw@shanghaitech.edu.cn}
\ead[URL]{https://faculty.sist.shanghaitech.edu.cn/faculty/tukw/}

\credit{Supervision, Writing – review \& editing, Methodology, Funding acquisition}

\cortext[cor1]{Principal corresponding author}
\cortext[cor2]{Corresponding author}


\begin{abstract}
Named entity recognition (NER) models often struggle with noisy inputs, such as those with spelling mistakes or errors generated by Optical Character Recognition processes, and learning a robust NER model is challenging. Existing robust NER models utilize both noisy text and its corresponding gold text for training, which is infeasible in many real-world applications in which gold text is not available. In this paper, we consider a more realistic setting in which only noisy text and its NER labels are available. We propose to retrieve relevant text of the noisy text from a knowledge corpus and use it to enhance the representation of the original noisy input. We design three retrieval methods: sparse retrieval based on lexicon similarity, dense retrieval based on semantic similarity, and self-retrieval based on task-specific text. After retrieving relevant text, we concatenate the retrieved text with the original noisy text and encode them with a transformer network, utilizing self-attention to enhance the contextual token representations of the noisy text using the retrieved text. We further employ a multi-view training framework that improves robust NER without retrieving text during inference. Experiments show that our retrieval-augmented model achieves significant improvements in various noisy NER settings.
\end{abstract}




\maketitle

\section{Introduction}
Named entity recognition (NER) \citep{chinchor1997muc} is an important task in natural language processing that has a variety of applications and serves as a foundation for several downstream tasks such as machine translation \citep{halek2011named}, intelligent question answering \citep{lee2007fine}, knowledge graphs \citep{thukral2023knowledge}, and event extraction \citep{zhang2022combining, edouard2017graph}. The performance of NER directly impacts the performance of these downstream tasks. However, current NER models are trained on clean text and pre-trained models used for NER are also based on clean text. In reality, input text may contain noises, e.g., spelling mistakes or errors from Optical Character Recognition (OCR) processes, which can lead to a drop in NER performance and propagate errors to downstream tasks. Therefore, robust NER that can handle both clean and noisy text is a valuable task in need of further research.

\begin{figure}
\centering
\subcaptionbox{existing robust NER models}{
\includegraphics[width=0.4\textwidth]{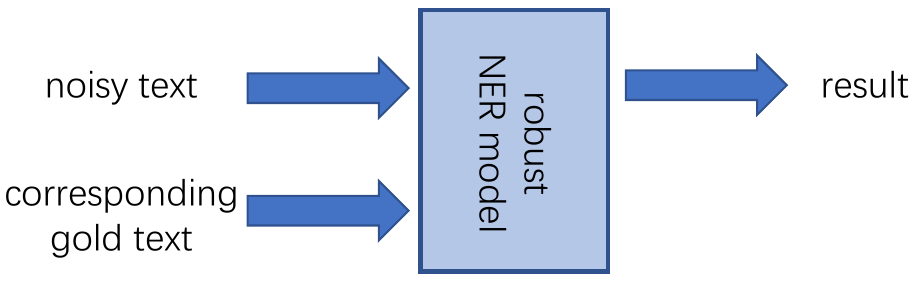}
}
\subcaptionbox{our robust NER model}{
\includegraphics[width=0.4\textwidth]{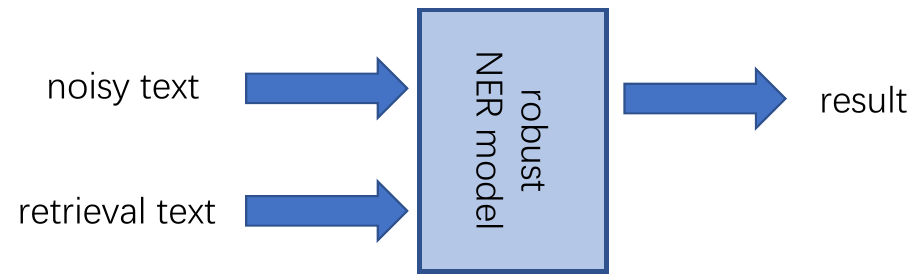}
}
\caption{A Comparative Illustration of the Divergence Between the Frameworks of Existing Robust Named Entity Recognition Models and Our Robust Named Entity Recognition Model} 
\label{Fig.0}
\end{figure}

Existing models \citep{namysl-etal-2020-nat, namysl-etal-2021-empirical} for robust NER rely on the use of both noisy text and its corresponding gold text for training. Specifically, they are trained to output similar results from noisy and gold texts. The NAT model \citep{namysl-etal-2020-nat} handles noisy text with spelling mistakes, which are token-level noises and do not change the length of sentences. It is trained to encourage similar predictions at each position from noisy and gold texts. However, it cannot handle errors generated by OCR which are sentence-level noises that may change the length of sentences. An extended model \citep{namysl-etal-2021-empirical} aligns a noisy sentence with OCR errors to the corresponding gold sentence at the token level using the Levenshtein Distance to ensure that the two sentences have the same length. Specifically, if a single word is recognized as multiple separate words by the OCR process, then these recognized words are treated as a cohesive unit aligned to the original word; if a single word is not recognized at all, then it is treated as a ``blank''. After the alignment process is complete, the lengths of the noisy and gold texts are equal and hence they can be inputted into the NAT model \citep{namysl-etal-2020-nat}.

Unlike prior studies, this paper addresses the challenge of robust NER in a more realistic setting where only noisy training text is available and its corresponding gold text is absent (as illustrated in Figure ~\ref{Fig.0}). Additionally, we assume that NER annotations are based on the noisy text rather than the gold text, so for texts with OCR errors, the number of labels is equal to the number of recognized words after OCR, which is again more realistic.

To tackle robust NER in this demanding setting, we draw inspiration from humans. Humans are capable of recognizing and annotating entities in noisy text by using domain knowledge retrieved from their memory to correct errors and resolve ambiguity.
Accordingly, we propose to solve robust NER in two steps. First, we retrieve relevant text of the noisy text from a background knowledge corpus. Second, we concatenate the noisy text with the retrieved text and feed them into a transformer. The self-attention mechanism of the transformer can enhance the contextual token representation of the noisy text, leveraging the information from the retrieved text. We then use the enhanced contextual token representation for the final NER prediction. Our method can be seen as an extension of retrieval-augmented NER \citep{wang-etal-2021-improving, wang-etal-2022-ita, Wang2022NamedEA}, but unlike previous work, we use text retrieval for the purpose of de-noising.

We use three retrieval methods in the first step, namely, sparse retrieval based on lexicon similarity, dense retrieval based on semantic similarity, and self-retrieval based on task-specific text.

\begin{itemize}
\item \textbf{Sparse retrieval based on lexicon similarity.}
This simple and effective method uses BM25 \citep{robertson1994some} to perform sparse retrieval based on string matching and retrieve relevant information from Wikipedia, a constantly evolving source of knowledge.

\item \textbf{Dense retrieval based on semantic similarity.}
BM25 requires at least one word from the query text to match. To overcome this limitation, we can use dense retrieval \citep{deerwester1990indexing}, which first encodes sentences into embeddings and then retrieves through K-Nearest Neighbors. To train the encoder, we employ a contrastive learning approach \citep{gao-etal-2021-simcse} using both clean text from Wikipedia and corresponding simulated noisy text as inputs.

\item \textbf{Self retrieval based on task-specific text.}
It is likely that a word may appear for multiple times in a noisy dataset in different forms: either its correct form or various noisy forms. Therefore, we can use noisy text as query to retrieve from the noisy dataset itself, hoping that the retrieved text may contain correct or less noisy forms of words in the query text. BERTScore \citep{zhangbertscore}, which uses contextual embeddings to represent tokens and calculates matching using cosine similarity, is more fine-grained than dense retrieval that directly uses sentence representation. However, BERTScore requires calculating the cosine similarity of each pair of token representations of two sentences and storing all contextual token embeddings, making it infeasible for use with Wikipedia.
\end{itemize}

While our method provides robust NER without relying on gold text during training, it requires retrieval from a background knowledge corpus. This may not be always possible or desirable, particularly in time-sensitive situations or online serving where retrieval could be too time-consuming. To address this issue, we use multi-view learning \citep{clark-etal-2018-semi,wang-etal-2021-improving} to train the NER model from two views, one being the original noisy text view and the other being the retrieval-based view that concatenates the noisy and retrieved texts. We encourage the two views to produce similar contextual embedding representations or predicted label distributions, thus improving the NER accuracy of the original noisy text view.

The contributions of this paper can be summarized as follows:

\begin{itemize}

\item We propose the first approach to robust NER with only noisy training text and without its corresponding gold text.

\item We empirically show that retrieval-augmentation can improve noisy named entity recognition with misspelling mistakes and OCR errors, and our model achieves significant improvements over the baseline in various experimental settings, i.e., with various levels of misspelling mistakes and types of OCR errors. 

\item We use three retrieval methods, specifically, self retrieval based on task-specific text, i.e., BERTScore to retrieve noisy dataset self, sparse retrieval based on lexicon similarity, i.e., BM25 to retrieve Wikipedia, and dense retrieval based on semantic similarity to retrieve Wikipedia. In the dense retrieval, we train the encode by a contrastive learning method using Wikipedia's clean text and its corresponding synthetic noisy text.

\end{itemize}

\section{Problem Definition}
\begin{figure}
\centering
\includegraphics[width=0.45\textwidth]{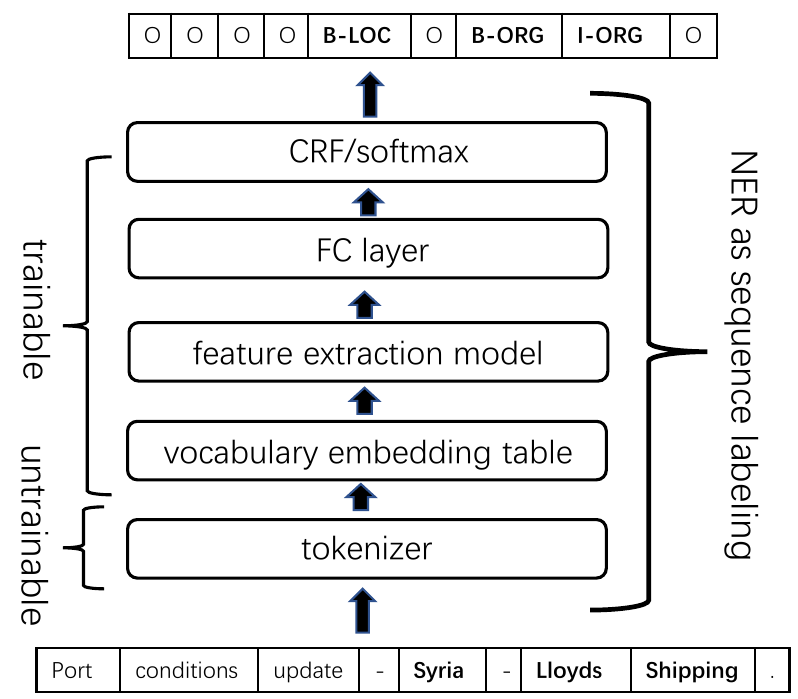}
\caption{NER as sequence labeling.}
\label{Fig.1}
\end{figure}

We model the named entity recognition (NER) task as a sequence labeling problem. Given a sentence $\boldsymbol{w}=\left\{x_1, \cdots, x_n\right\}$ with $n$ tokens, sequence labeling aims to predict a label sequence $\boldsymbol{y}=\left\{y_1, \cdots, y_n\right\}$ for each position. The BIO tagging scheme, which stands for Beginning, Inside and Outside subtags, is often used. A typical architecture of sequence labeling for NER is shown in Figure~\ref{Fig.1}.

To reduce the vocabulary size and better handle similar words, natural language models often use word segmentation \citep{saffran1996word}, i.e., tokenizers. 
Sequence labeling begins with tokenizing the sentence into smaller subwords, which are then transformed into token embeddings using a vocabulary embedding table. These token embeddings are then inputted into a feature extraction model, which outputs contextual token embeddings based on interactions among tokens of different positions. The feature extraction model can be a pre-trained transformer, such as BERT \citep{devlin-etal-2019-bert} or RoBERTa \citep{liu2019roberta}, which has powerful text representation ability due to the self-attention mechanism and pre-training on large corpora using mask language modeling.

To account for the fact that a single token can be tokenized into multiple subtokens, a pooling operation is used to merge the contextual token embeddings at consecutive positions to create token-level embeddings, such as first pooling, i.e., using the embedding of the first sub-token. The pooled contextual token embeddings are then converted to non-normalized logits in each position through a fully connected layer. A softmax layer can be added to get normalized logits for the label in each position, but in practice, a Conditional Random Fields (CRF) layer is often used which effectively considers relationship between consecutive labels to enforce constraints, such as ensuring that there is a Beginning tag before an Inside tag.

The vocabulary embedding table, the feature extraction model, and the CRF layer can be jointly trained using backpropagation, while the tokenizer is not trained and the size and words of the vocabulary embedding table remain unchanged.

When the input text is noisy, the process of tokenization may produce different results compared with that from the corresponding clean text. This can lead to errors that will propagate through to the final predicted result. There are many kinds of noise in the text and two of the most important ones are spelling mistakes and errors generated by Optical Character Recognition (OCR). Spelling mistakes alter individual letters within words (including replacing, adding and deleting), resulting in token-level noises, but they do not change the number of words in a sentence. Words with spelling mistakes are often tokenized into single characters or marked as unknown words after tokenization. On the other hand, OCR-generated errors may change the number of words of a sentence, resulting in sentence-level noises. The OCR process recognizes words from images of texts and hence the wrongly recognized words are often highly similar to the original ones, so they may be tokenized into fragments of the original words. The difference of tokenization between spelling mistakes and OCR errors can be demonstrated through the BERTScore experiment (Section~\ref{Noise Induction}), where the BERTScore \citep{zhangbertscore} for text with spelling mistakes is low, while that for text with OCR errors is relatively high, indicating that text with OCR errors is quite similar to its corresponding original text. The correlation between the similarity of two sentences and their BERTScores is direct, with increased similarity leading to a higher BERTScore score.

Unlike previous work, we assume that during training, we only have noisy training text and do not know its corresponding gold text. Furthermore, we assume that NER annotations are derived from the noisy text, as opposed to the gold text. In this setting, the number of labels for OCR-erroneous texts is equivalent to the number of words recognized through OCR, which is more practical and realistic.

\section{Method}
\begin{figure}
\centering
\includegraphics[width=0.48\textwidth]{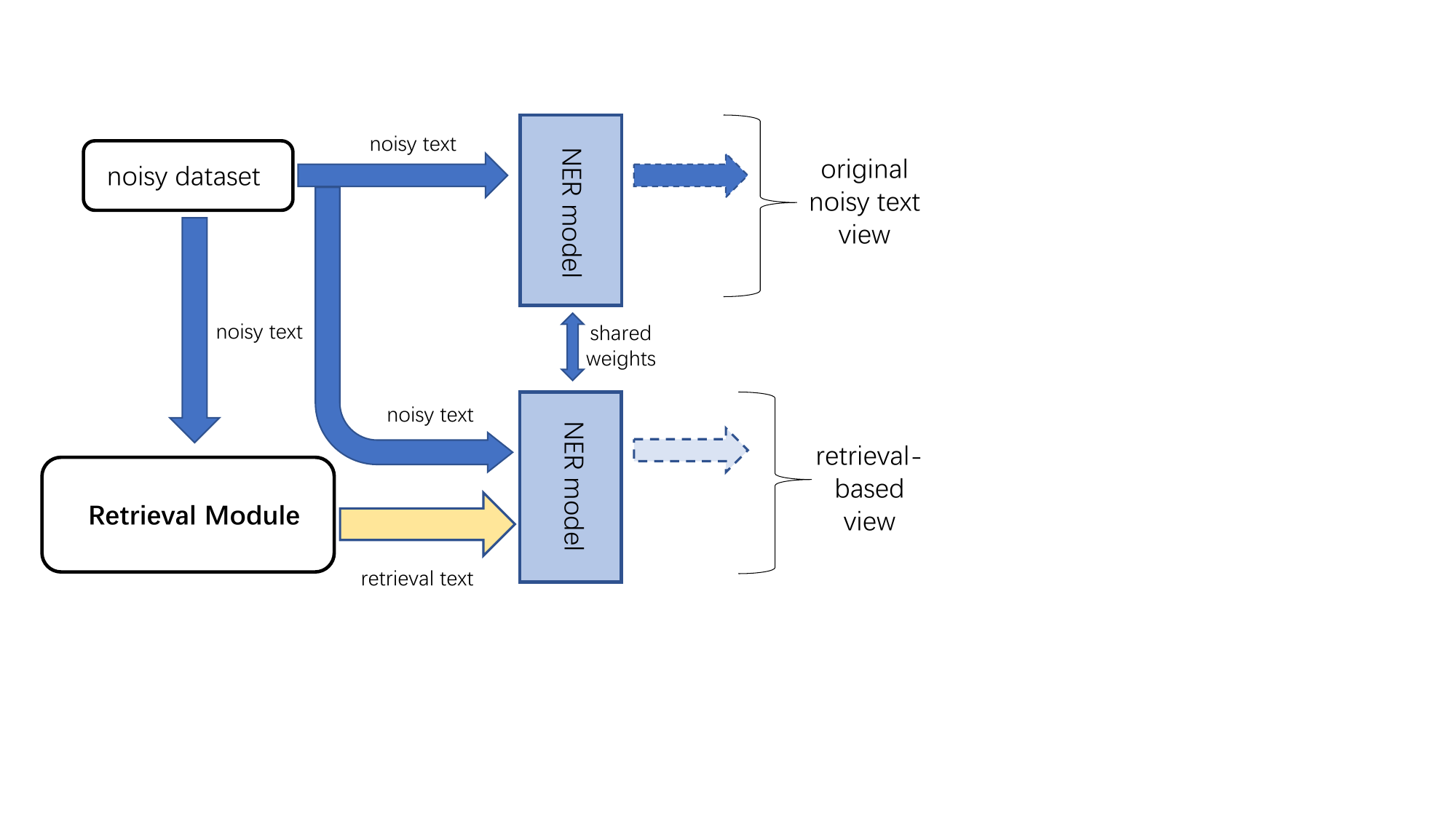}
\caption{Our model architecture}
\label{Fig.2}
\end{figure}

Our proposed framework consists of two main components: a Retrieval Module and a NER model with multi-view learning, as depicted in Figure~\ref{Fig.2}.
The Retrieval Module is used to obtain text related to the input text from a background knowledge corpus. The NER model is a transformer-based model with two views.
In the retrieval-based view, we concatenate the original noisy text with the retrieved text and input them into the transformer-based model, hoping to enhance the contextual token representation of the noisy text leveraging the information from the retrieved text. In the original noisy text view, the noisy text is directly inputted into the NER model without retrieval. We employ multi-view learning to encourage two views to produce similar contextual embeddings or predicted label distributions, ultimately improving the accuracy of the original noisy text view, which can be used when retrieval is impossible or undesirable.

\subsection{Retrieval Module}
We design three retrieval methods: sparse retrieval based on lexicon similarity , dense retrieval based on semantic similarity, and self retrieval based on task-specific text.

\subsubsection{Sparse retrieval based on lexicon similarity}
Wikipedia \footnote{https://www.wikipedia.org/}, the largest encyclopedia in the world, boasts a vast collection of clean and well-organized text on a wide range of topics and domains. The simplest method for retrieving information from Wikipedia is through the BM25 algorithm \citep{robertson1994some}, a string matching-based approach.

\subsubsection{Dense retrieval based on semantic similarity}
The BM25 algorithm requires that retrieved strings contain at least one word from the input query, making it less effective for retrieving results from noisy text containing errors. Fuzzy retrieval based on Levenshtein distance may be used instead. However, due to the vast size of Wikipedia, the word with the smallest Levenshtein distance to the noisy text may not necessarily be the correct word, resulting in sub-optimal results. To address this, we apply a dense retrieval method, which encodes sentences into embeddings and then retrieves text through K-Nearest Neighbors. To obtain the encoder, we utilize contrastive learning \citep{gao-etal-2021-simcse} to fine-tune a pre-trained model using clean text from Wikipedia and its noisy version with simulated corruptions. We use the InfoNCE loss \citep{oord2018representation} as the training loss:
\begin{equation}
\begin{aligned}
\mathcal{L}_i=-\log \frac{e^{\operatorname{sim}\left(\mathbf{h}_i^{noisy}, \mathbf{h}_i^{gold}\right) / \tau}}{\sum_{j=1}^Ne^{\operatorname{sim}\left(\mathbf{h}_i^{noisy}, \mathbf{h}_j^{gold}\right) / \tau}}
\end{aligned}
\end{equation}
where $\mathbf{h}_i^{noisy}$ is the embedding of noisy text, $\mathbf{h}_i^{gold}$ is the embedding of the corresponding gold text, $\tau$ is the temperature hyperparameter, $\operatorname{sim}\left(\cdot, \cdot\right)$ is the cosine similarity, and $N$ is the batchsize.
We select the best encoder model by measuring $\emph{{recall@}k}$ of using simulated noisy text of Wikipedia to retrieve the corresponding gold text. $\emph{{recall@}k}$ is measured by comparing the top $k$ retrieved results with the gold text and calculating the proportion of times the gold text is included in the top $k$ results, as shown by the following equation:
\begin{equation}
\begin{aligned}
\emph{{recall@}k}=\frac{\sum_{i=1}^n \mathbb{I}\left[\textup{gold}_i \in \mathbb{R}_{topk}(\textup{noisy}_{i}) \right]}{n}
\end{aligned}
\end{equation}
where $\textup{noisy}_{i}$ is the $i$-$th$ simulated noisy text of Wikipedia, $\textup{gold}_i$ is the corresponding clean gold text, $\mathbb{R}_{topk}(\textup{x})$ is the top-k results of using $\textup{x}$ to retrieve clean texts from Wikipedia, $\mathbb{I}\left[x\right]$ is equal to $1$ if $x$ is true and $0$ otherwise, and $n$ is the number of simulated noisy texts.

An encoder trained in this way could produce similar embeddings for noisy text and its corresponding gold text. Subsequently, we utilize the K-nearest neighbors (knn) algorithm to retrieve Wikipedia sentences.

\subsubsection{Self retrieval based on task-specific text}
The dense retrieval method encodes an entire sentence into a single embedding. A more fine-grained retrieval method is to use BERTScore \citep{zhangbertscore}, which calculates the average maximum cosine similarity between the contextual token embeddings of the input sentence and the retrieved sentence, which is then used to determine the precision (P) and recall (R). We use F1-score (harmonic mean of P and R) to rank the retrieved sentences.
However, BERTScore has high computational complexity and a large memory footprint, making it difficult to use on large datasets like Wikipedia. Therefore, we apply it to retrieve text from the training dataset itself, which is small and contains sentences of the same domain as the query sentence.
For the dataset self, it is smaller and from the same domain, where the same words may have the same semantics, appear multiple times and may or may not be changed by a noising process. The BERTScore's F1-score can be used to retrieve and rank the noisy dataset itself. Since our experiment is inductive, with test instances being independent of each other, we retrieve from the same training set for all sets: training, validation, and test.

\subsection{NER model with multi-view learning}
For the NER model, we use a pre-trained transformer-based model and a CRF layer as the backbone. Our method employs two views: the original noisy text view and the retrieval-based view. Both views share the same parameters in their pre-trained transformer-based models and CRF layers. The retrieval-based view combines the noisy text with the corresponding retrieved text from the Retrieval Module to form a concatenated input $[\boldsymbol{x}, [\boldsymbol{X}], \tilde{\boldsymbol{x}}]$, where $\boldsymbol{x}$ is the noisy text, $\tilde{\boldsymbol{x}}$ is the corresponding retrieved text, and $[\boldsymbol{X}]$ is a special token dependent on the pre-trained transformer-based model used by NER (e.g., ``[SEP]'' in BERT and ``</s>'' in RoBERTa). The transformer-based embedding model, utilizing its self-attention mechanism, enhances the contextual token representations of the noisy text by the retrieval text, ultimately leading to improved NER performance. Note that in the retrieval-based view, we only input the embeddings of the original noisy text into the CRF layer, discarding the embedding of the retrieval text since we do not need to predict their labels.

As discussed earlier, retrieval is slow and may not be suitable for time-sensitive scenarios. Therefore, we use multi-view learning to improve the original noisy text view with the help from the retrieval-based view.
For the original noisy text view, denote the contextual token representations after the pre-trained transformer as $\left\{\boldsymbol{r}_1, \cdots, \boldsymbol{r}_n\right\}$. After the CRF layer, the label distribution is:
\begin{equation}
\begin{aligned}
p_\theta(\boldsymbol{y}|\boldsymbol{x})=\frac{\prod_{i=1}^n \psi\left(y_{i-1}, y_i, \boldsymbol{r}_i\right)}{\sum_{\boldsymbol{y}^{\prime} \in \mathcal{Y}(\boldsymbol{x})} \prod_{i=1}^n \psi\left(y_{i-1}^{\prime}, y_i^{\prime}, \boldsymbol{r}_i\right)}
\end{aligned}
\end{equation}
where $\theta$ is the model parameters, $\psi$ is the potential function, and $\mathcal{Y}(\boldsymbol{x})$ is the set of all possible label sequences given the sentence $\boldsymbol{x}$. Note that the denominator can be computed using the classic forward algorithm.
Denote the gold label sequence as $\hat{\boldsymbol{y}}$. The negative log-likelihood loss for the original noisy text view is:
\begin{equation}
\begin{aligned}
\mathcal{L}_{\mathrm{text}}(\theta)=-\log p_\theta(\hat{\boldsymbol{y}}|\boldsymbol{x})
\end{aligned}
\end{equation}

Similarly, the negative log-likelihood loss of the retrieval-based view is
\begin{equation}
\begin{aligned}
\mathcal{L}_{\mathrm{retrieval}}(\theta)=-\log p_\theta(\hat{\boldsymbol{y}}|\boldsymbol{x,\tilde{\boldsymbol{x}}})
\end{aligned}
\end{equation}

To jointly train the two views by multi-view learning, we use L2 distance or KL distance to reduce the gap between the two views. Specifically, the L2 distance is the distance between the token representations of the two views after the transformer-based embedding model:
\begin{equation}
\begin{aligned}
\mathcal{L}_{\mathrm{L_2}}(\theta)=\sum_{i=1}^n\left\|\boldsymbol{r}_i^{\prime}-\boldsymbol{r}_i\right\|_2^2
\end{aligned}
\end{equation}
The KL distance is the distance between the predicted label distributions of the two views after the CRF layer:
\begin{align}
\mathcal{L}_{\mathrm{KL}}(\theta)&=\operatorname{KL}\left(p_\theta(\boldsymbol{y}|\boldsymbol{x}, \tilde{\boldsymbol{x}}) \| p_\theta(\boldsymbol{y}|\boldsymbol{x})\right) \\
&= \operatorname{CE}\left(p_\theta(\boldsymbol{y}|\boldsymbol{x}, \tilde{\boldsymbol{x}}), p_\theta(\boldsymbol{y}|\boldsymbol{x})\right) - \operatorname{H}\left(p_\theta(\boldsymbol{y}|\boldsymbol{x}, \tilde{\boldsymbol{x}})\right) \\
&\triangleq \operatorname{CE}\left(p_\theta(\boldsymbol{y}|\boldsymbol{x}, \tilde{\boldsymbol{x}}), p_\theta(\boldsymbol{y}|\boldsymbol{x})\right) \label{eq:third_line}\\
&= -\sum_{\boldsymbol{y} \in \mathcal{Y}(\boldsymbol{x})} p_\theta(\boldsymbol{y}|\boldsymbol{x}, \tilde{\boldsymbol{x}}) \log p_\theta(\boldsymbol{y}|\boldsymbol{x}) \\
&\circeq -\sum_{i=1}^n q_\theta\left(y_i|\boldsymbol{x}, \tilde{\boldsymbol{x}}\right) \log q_\theta\left(y_i|\boldsymbol{x}\right) \label{eq:fifth_line}
\end{align}
where $\operatorname{KL}\left(\cdot\|\cdot\right)$ denotes the Kullback-Leibler divergence, $\operatorname{CE}\left(\cdot\|\cdot\right)$ represents the Cross-Entropy, $\operatorname{H}\left(\cdot\right)$ signifies the Entropy, and $q_\theta\left(y_k|*\right)$ is proportional to the product of the forward score and backward score generated by the classic forward-backward algorithm.

Note that the retrieval-based view has additional information of retrieval text so we does not necessitate parameter adjustment through back-propagation. Consequently, $\operatorname{H}\left(p_\theta(\boldsymbol{y}|\boldsymbol{x}, \tilde{\boldsymbol{x}})\right)$ remains constant and can be disregarded, leading to Equation \eqref{eq:third_line} $\triangleq$.

Due to the exponential size of $\mathcal{Y}(\boldsymbol{x})$, we approximate the Cross-Entropy over the label sequence with the summation of position-wise Cross-Entropy, as shown in Equation \eqref{eq:fifth_line} $\circeq$.

In the training phase, we simultaneously train the two views by optimizing the sum of the loss functions:
\begin{equation}
\begin{aligned}
\mathcal{L}(\theta)=\mathcal{L}_{\mathrm{text}}(\theta)+\mathcal{L}_{\mathrm{retrieval}}(\theta)+\mathcal{L}_{\mathrm{MV}}(\theta)
\end{aligned}
\end{equation}
where $\mathcal{L}_{\mathrm{MV}}(\theta)$ is either $\mathcal{L}_{\mathrm{L_2}}(\theta)$ or $\mathcal{L}_{\mathrm{KL}}(\theta)$.

During the inference stage, the original noisy text view is fed with the noisy text input, while the retrieval-based view receives a combination of the noisy text and its corresponding retrieved text. The two views are evaluated separately via entity-level F1 scores.

\section{Experiment}

\subsection{Dataset Construction}
\label{Noise Induction}
In light of the absence of publicly accessible noisy datasets specifically designed for Named Entity Recognition (NER) tasks – with an emphasis on noisy text rather than noisy labels – we have undertaken the development of our own noisy datasets. These custom datasets are constructed using two established NER corpora as a foundation: the WNUT-17 dataset \citep{derczynski-etal-2017-results} collected from social media and the CoNLL-03 English dataset \citep{tjong-kim-sang-de-meulder-2003-introduction} derived from news.

We induce two types of noise into clean text, i.e., spelling mistakes and errors caused by Optical Character Recognition (OCR). To simulate spelling mistakes, we employ the NAT model proposed by Namysl et al. (2020) \citep{namysl-etal-2020-nat}. This model uses a character confusion matrix to generate errors, where the probabilities of insertion, deletion, and substitution are each equal to $p/3$, given an overall noise level of $p$. The alphabet $\Sigma$, which excludes the symbol $\varepsilon$, comprises all letters present in the original corpus. In applying this method, we first insert the $\varepsilon$ symbol between every pair of adjacent letters in the original text, as well as before the first letter and after the last letter. Next, we utilize the character confusion matrix to create the modified sequence. Specifically, insertions transform $\varepsilon$ into a character from $\Sigma$, deletions change the original character to $\varepsilon$, and substitutions replace the original character with another distinct character from $\Sigma$. Finally, we remove the $\varepsilon$ symbols to produce the resulting noisy text containing spelling errors.

In order to simulate errors introduced by OCR, we make use of the Text Recognition Data Generator (TRDG) package\footnote{https://github.com/Belval/TextRecognitionDataGenerator} to convert each sentence into an image, employing 90 distinct fonts. Subsequently, we introduce random distortions to the sentences within the images or incorporate background noise. To extract potentially noisy text from these images, we employ the Tesseract-OCR engine\footnote{https://github.com/tesseract-ocr/tesseract}.
Considering that OCR may alter sentence length, we align the noisy OCR sentences with their respective clean counterparts using the Levenshtein Distance Algorithm, consistent with prior work \citep{namysl-etal-2021-empirical}. In our approach, the number of labels corresponds to the total number of words recognized following the OCR process. This differs from previous research \citep{namysl-etal-2021-empirical}, where the number of labels equaled the number of original words, which does not reflect real-world scenarios.

We use three different noise levels of spelling mistakes, i.e., $p=0.1,0.2,0.3$, and four kinds of OCR noise, i.e., whether to combine geometrical distortions for sentences and whether to induce pixel-level noise to the background, whose token error rate can be seen in Table~\ref{four ways of OCR noise}. The token error rate is calculated as the ratio of the Levenshtein Distance between the noisy and original sentences to the total length of the original sentences. The noisy level of the noisy sentences and its Levenshtein Distance are positively correlated: as the Levenshtein Distance increases, the noisy level also increases.

\begin{table}[tb]
\centering
\caption{Four ways of OCR noise.}
\resizebox{0.8\linewidth}{!}{
\begin{tabular}{ccccc}
\hline
\rule{0pt}{12pt}
\makecell*[c]{OCR\\type}&\makecell*[c]{distorted\\sentences}&\makecell*[c]{noisy\\backdrop}&\makecell*[c]{WNUT 17}& \makecell*[c]{CoNLL 03}\\
\hline
OCR\_1 & \ding{55} & \ding{55} & 2.53\% & 1.81\% \\
OCR\_2 & \ding{51} & \ding{55} & 8.07\% & 5.37\%\\
OCR\_3 & \ding{55} & \ding{51} & 13.62\% & 10.84\%\\
OCR\_4 & \ding{51} & \ding{51} & 26.41\% & 21.23\%\\
\hline
\end{tabular}}
\label{four ways of OCR noise}
\end{table}

To evaluate the differences between the two noise-induced methods, we conducted an experiment using BERTScore to measure the similarity between the noisy text and the corresponding gold text as shown in Table~\ref{BERTScore experiment}, where $typos\_p$ in the dataset is the noisy dataset generated using spelling mistakes with the noise level $p$, $OCR\_1/OCR\_2/OCR\_3/OCR\_4$ denote the noisy dataset induced by four kinds of OCR noise. We use two pre-trained transformer models in the BERTScore experiment, i.e., DeBERTa-XLarge-MNLI and XLM-RoBERTa-large. The DeBERTa-XLarge-MNLI model \citep{he2021deberta} is the best model for BERTScore and the XLM-RoBERTa-large model \citep{conneau-etal-2020-unsupervised} is used in our NER model. The BERTScore of sentences with spelling mistakes is really low while that with OCR noise is very high, which indicates that spelling mistakes and OCR noise are very different.

\begin{table}[tb]
\centering
\caption{BERTScore between noisy text and the corresponding gold text.}
\resizebox{1.0\linewidth}{!}{
\begin{tabular}{c|cc|cc}
\hline
\rule{0pt}{12pt}
 & \multicolumn{2}{c|}{\makecell*[c]{DeBERTa-XLarge-MNLI}} & \multicolumn{2}{c}{\makecell*[c]{XLM-RoBERTa-large}} \\
\hline
dataset&WNUT 17&CoNLL 03&WNUT 17&CoNLL 03\\
\hline
typos\_0.1 & $0.791\pm0.092$& $0.790\pm0.110$& $0.923\pm0.029$& $0.923\pm0.037$\\
typos\_0.2 & $0.657\pm0.100$& $0.660\pm0.125$& $0.880\pm0.030$& $0.883\pm0.040$\\
typos\_0.3 & $0.561\pm0.097$& $0.571\pm0.129$& $0.849\pm0.030$& $0.856\pm0.042$\\
\hline
OCR\_1 & $0.964\pm0.048$& $0.977\pm0.051$& $0.985\pm0.019$& $0.992\pm0.018$\\
OCR\_2 & $0.878\pm0.094$& $0.920\pm0.091$& $0.955\pm0.032$& $0.971\pm0.032$\\
OCR\_3 & $0.829\pm0.184$& $0.875\pm0.172$& $0.939\pm0.061$& $0.956\pm0.058$\\
OCR\_4 & $0.696\pm0.202$& $0.779\pm0.209$& $0.896\pm0.065$& $0.924\pm0.067$\\
\hline
\end{tabular}}
\label{BERTScore experiment}
\end{table}

\begin{table*}[tb]
    \setlength{\tabcolsep}{3pt}
    \centering
    \caption{NER performance with spelling mistakes}
    \label{the NER performance of spelling mistakes}
    \subcaptionbox{WNUT-17}
    {
        \resizebox{0.483\linewidth}{!}{
        \begin{tabular}{c|cc|cc|cc|cc}
            \hline
            \rule{0pt}{12pt}
            \multirow{2}*{noise level} &
            \multicolumn{2}{c|}{0} & \multicolumn{2}{c|}{0.1} & \multicolumn{2}{c|}{0.2} & \multicolumn{2}{c}{0.3}\\
            \cline{2-9}
            \rule{0pt}{12pt}
            & OV & RV & OV & RV & OV & RV & OV & RV \\
            \hline
            \multicolumn{1}{l|}{\emph{Baselines:}} & & & & & & & & \\
            \multicolumn{1}{l|}{\emph{noisy text only}} & & & & & & & & \\
            w/o context & 58.21& -& 46.26& -& 33.45& -& 24.72& -\\
            \hline
            \multicolumn{1}{l|}{\emph{correct}} & & & & & & & & \\
            \multicolumn{1}{l|}{\emph{w/o context}} & & & & & & & & \\
            natas & -& -& 37.27& -& 27.04& -& 18.20& -\\
            hunspell & -& -& 44.46& -& 32.03& -& 23.62& -\\
            natas+check & -& -& 44.99& -& 33.71$^{\dagger}$& -& 22.98& -\\
            hunspell+check & -& -& 50.85$^{\dagger}$& -& 40.27$^{\dagger}$& -& 30.52$^{\dagger}$& -\\
            \hline
            \multicolumn{1}{l|}{\emph{with gold text}} & & & & & & & & \\
            w/ gold full & -& -& -& 55.33$^{\dagger}$& -& 53.25$^{\dagger}$& -& 52.93$^{\dagger}$\\
            w/ gold KL & -& -& 46.57$^{\dagger}$& 53.97$^{\dagger}$& 36.54$^{\dagger}$& 53.60$^{\dagger}$& 29.03$^{\dagger}$& 53.90$^{\dagger}$\\
            w/ gold L2 & -& -& 46.80$^{\dagger}$& 54.84$^{\dagger}$& 36.44$^{\dagger}$& 53.28$^{\dagger}$& 29.04$^{\dagger}$& 53.93$^{\dagger}$\\
            \hline
            \multicolumn{1}{l|}{\emph{self-retrieval}} & & & & & & & & \\
            w/ self full & -& 58.90$^{\dagger}$& -& 47.08$^{\dagger}$& -& 35.68$^{\dagger}$& -& 26.21$^{\dagger}$\\
            w/ self KL & 59.59$^{\dagger}$& 59.87$^{\dagger}$& 47.08$^{\dagger}$& 47.67$^{\dagger}$& 35.83$^{\dagger}$& 36.84$^{\dagger}$& 26.45$^{\dagger}$& 28.47$^{\dagger}$\\
            w/ self L2 & 59.02$^{\dagger}$& 59.21$^{\dagger}$& 47.06$^{\dagger}$& 48.29$^{\dagger}$& 35.94$^{\dagger}$& 36.36$^{\dagger}$& 26.59$^{\dagger}$& 28.87$^{\dagger}$\\
            \hline
            \multicolumn{1}{l|}{\emph{BM25-retrieval}} & & & & & & & & \\
            w/ para full & -& 59.29$^{\dagger}$& -& 47.85$^{\dagger}$& -& 33.28$^{\dagger}$& -& 26.05$^{\dagger}$\\
            w/ sent-link full & -& 59.88$^{\dagger}$& -& 46.57$^{\dagger}$& -& 33.37$^{\dagger}$& -& 26.56$^{\dagger}$\\
            w/ sent full & -& 59.15$^{\dagger}$& -& 46.56$^{\dagger}$& -& 34.42$^{\dagger}$& -& 23.43$^{\dagger}$\\
            w/ para KL & 59.39$^{\dagger}$& 60.02$^{\dagger}$& 47.36$^{\dagger}$& 48.04$^{\dagger}$& 35.73$^{\dagger}$& 35.66$^{\dagger}$& 27.01$^{\dagger}$& 27.21$^{\dagger}$\\
            w/ sent-link KL & 59.54$^{\dagger}$& 60.61$^{\dagger}$& 47.34$^{\dagger}$& 47.88$^{\dagger}$& 35.16$^{\dagger}$& 34.82$^{\dagger}$& 26.45$^{\dagger}$& 26.35$^{\dagger}$\\
            w/ sent KL & 59.21$^{\dagger}$& 59.65$^{\dagger}$& 48.47$^{\dagger}$& 48.13$^{\dagger}$& 35.71$^{\dagger}$& 35.16$^{\dagger}$& 27.14$^{\dagger}$& 27.12$^{\dagger}$\\
            w/ para L2 & 59.94$^{\dagger}$& 59.92$^{\dagger}$& 46.98$^{\dagger}$& 47.79$^{\dagger}$& 35.69$^{\dagger}$& 35.85$^{\dagger}$& 27.45$^{\dagger}$& 27.56$^{\dagger}$\\
            w/ sent-link L2 & 59.02$^{\dagger}$& 59.05$^{\dagger}$& 47.28$^{\dagger}$& 48.19$^{\dagger}$& 34.97$^{\dagger}$& 35.14$^{\dagger}$& 26.77$^{\dagger}$& 27.24$^{\dagger}$\\
            w/ sent L2 & 59.10$^{\dagger}$& 59.39$^{\dagger}$& 47.36$^{\dagger}$& 46.99$^{\dagger}$& 35.22$^{\dagger}$& 34.77$^{\dagger}$& 26.72$^{\dagger}$& 26.55$^{\dagger}$\\
            \hline
            \multicolumn{1}{l|}{\emph{dense-retrieval}} & & & & & & & & \\
            w/ para full & -& 58.94$^{\dagger}$& -& 46.59$^{\dagger}$& -& 35.07$^{\dagger}$& -& 26.12$^{\dagger}$\\
            w/ sent-link full & -& 58.87$^{\dagger}$& -& 47.20$^{\dagger}$& -& 34.99$^{\dagger}$& -& 27.12$^{\dagger}$\\
            w/ sent full & -& 58.63$^{\dagger}$& -& 46.89$^{\dagger}$& -& 34.41$^{\dagger}$& -& 26.54$^{\dagger}$\\
            w/ para KL & 59.25$^{\dagger}$& 59.59$^{\dagger}$& 46.59$^{\dagger}$& 46.42$^{\dagger}$& 35.80$^{\dagger}$& 36.35$^{\dagger}$& 26.53$^{\dagger}$& 26.86$^{\dagger}$\\
            w/ sent-link KL & 59.26$^{\dagger}$& 59.08$^{\dagger}$& 47.98$^{\dagger}$& 47.85$^{\dagger}$& 35.16$^{\dagger}$& 35.89$^{\dagger}$& 26.11$^{\dagger}$& 26.79$^{\dagger}$\\
            w/ sent KL & 59.20$^{\dagger}$& 58.96$^{\dagger}$& 46.61$^{\dagger}$& 46.96$^{\dagger}$& 35.57$^{\dagger}$& 35.09$^{\dagger}$& 26.14$^{\dagger}$& 27.16$^{\dagger}$\\
            w/ para L2 & 59.03$^{\dagger}$& 59.01$^{\dagger}$& 47.61$^{\dagger}$& 46.56$^{\dagger}$& 37.09$^{\dagger}$& 37.07$^{\dagger}$& 25.96$^{\dagger}$& 27.37$^{\dagger}$\\
            w/ sent-link L2 & 59.41$^{\dagger}$& 59.84$^{\dagger}$& 47.15$^{\dagger}$& 46.14$^{\dagger}$& 35.26$^{\dagger}$& 35.91$^{\dagger}$& 26.26$^{\dagger}$& 27.17$^{\dagger}$\\
            w/ sent L2 & 59.11$^{\dagger}$& 59.08$^{\dagger}$& 46.84$^{\dagger}$& 47.02$^{\dagger}$& 35.19$^{\dagger}$& 35.47$^{\dagger}$& 25.94$^{\dagger}$& 27.14$^{\dagger}$\\
            \hline
        \end{tabular} }
    }
    \hfill
    \subcaptionbox{CoNLL-03}
    {
        \resizebox{0.483\linewidth}{!}{
        \begin{tabular}{c|cc|cc|cc|cc}
            \hline
            \rule{0pt}{12pt}
            \multirow{2}*{noise level} &
            \multicolumn{2}{c|}{0} & \multicolumn{2}{c|}{0.1} & \multicolumn{2}{c|}{0.2} & \multicolumn{2}{c}{0.3}\\
            \cline{2-9}
            \rule{0pt}{12pt}
            & OV & RV & OV & RV & OV & RV & OV & RV \\
            \hline
            \multicolumn{1}{l|}{\emph{Baselines: }} & & & & & & & & \\
            \multicolumn{1}{l|}{\emph{noisy text only}} & & & & & & & & \\
            w/o context & 92.98& -& 88.82& -& 84.37& -& 79.53& -\\
            \hline
            \multicolumn{1}{l|}{\emph{correct}} & & & & & & & & \\
            \multicolumn{1}{l|}{\emph{w/o context }} & & & & & & & & \\
            natas & -& -& 84.06& -& 79.30& -& 73.52& -\\
            hunspell & -& -& 87.50& -& 82.17& -& 75.68& -\\
            natas+check & -& -& 88.81& -& 84.79$^{\dagger}$& -& 79.85$^{\dagger}$& -\\
            hunspell+check & -& -& 90.15$^{\dagger}$& -& 86.84$^{\dagger}$& -& 82.55$^{\dagger}$& -\\
            \hline
            \multicolumn{1}{l|}{\emph{with gold text}} & & & & & & & & \\
            w/ gold full & -& -& -& 92.53$^{\dagger}$& -& 92.77$^{\dagger}$& -& 92.84$^{\dagger}$\\
            w/ gold KL & -& -& 89.08$^{\dagger}$& 92.57$^{\dagger}$& 84.71$^{\dagger}$& 92.52$^{\dagger}$& 79.85$^{\dagger}$& 92.79$^{\dagger}$\\
            w/ gold L2 & -& -& 89.01$^{\dagger}$& 92.75$^{\dagger}$& 84.84$^{\dagger}$& 92.56$^{\dagger}$& 79.83$^{\dagger}$& 92.65$^{\dagger}$\\
            \hline
            \multicolumn{1}{l|}{\emph{self-retrieval}} & & & & & & & & \\
            w/ self full & -& 93.10$^{\dagger}$& -& 89.26$^{\dagger}$& -& 85.00$^{\dagger}$& -& 79.87$^{\dagger}$\\
            w/ self KL & 93.21$^{\dagger}$& 93.46$^{\dagger}$& 89.06$^{\dagger}$& 89.58$^{\dagger}$& 85.05$^{\dagger}$& 85.67$^{\dagger}$& 79.90$^{\dagger}$& 80.54$^{\dagger}$\\
            w/ self L2 & 93.13$^{\dagger}$& 93.54$^{\dagger}$& 89.14$^{\dagger}$& 89.61$^{\dagger}$& 85.12$^{\dagger}$& 85.81$^{\dagger}$& 79.96$^{\dagger}$& 80.54$^{\dagger}$\\
            \hline
            \multicolumn{1}{l|}{\emph{BM25-retrieval}} & & & & & & & & \\
            w/ para full & -& 93.33$^{\dagger}$& -& 89.22$^{\dagger}$& -& 84.97$^{\dagger}$& -& 79.24\\
            w/ sent-link full & -& 93.27$^{\dagger}$& -& 89.13$^{\dagger}$& -& 85.13$^{\dagger}$& -& 79.47\\
            w/ sent full & -& 93.33$^{\dagger}$& -& 89.17$^{\dagger}$& -& 84.88$^{\dagger}$& -& 79.18\\
            w/ para KL & 93.07$^{\dagger}$& 93.26$^{\dagger}$& 89.10$^{\dagger}$& 89.26$^{\dagger}$& 85.01$^{\dagger}$& 85.46$^{\dagger}$& 79.93$^{\dagger}$& 80.05$^{\dagger}$\\
            w/ sent-link KL & 93.05$^{\dagger}$& 93.41$^{\dagger}$& 89.01$^{\dagger}$& 89.42$^{\dagger}$& 85.09$^{\dagger}$& 85.40$^{\dagger}$& 79.81$^{\dagger}$& 80.05$^{\dagger}$\\
            w/ sent KL & 93.01$^{\dagger}$& 93.38$^{\dagger}$& 89.05$^{\dagger}$& 89.62$^{\dagger}$& 85.17$^{\dagger}$& 85.48$^{\dagger}$& 79.80$^{\dagger}$& 79.90$^{\dagger}$\\
            w/ para L2 & 93.04$^{\dagger}$& 93.26$^{\dagger}$& 89.02$^{\dagger}$& 89.39$^{\dagger}$& 84.98$^{\dagger}$& 85.40$^{\dagger}$& 79.95$^{\dagger}$& 75.25$^{\dagger}$\\
            w/ sent-link L2 & 93.12$^{\dagger}$& 93.50$^{\dagger}$& 89.05$^{\dagger}$& 89.43$^{\dagger}$& 84.84$^{\dagger}$& 85.33$^{\dagger}$& 80.05$^{\dagger}$& 80.14$^{\dagger}$\\
            w/ sent L2 & 93.05$^{\dagger}$& 93.29$^{\dagger}$& 89.05$^{\dagger}$& 89.48$^{\dagger}$& 85.17$^{\dagger}$& 85.49$^{\dagger}$& 79.82$^{\dagger}$& 80.05$^{\dagger}$\\
            \hline
            \multicolumn{1}{l|}{\emph{dense-retrieval}} & & & & & & & & \\
            w/ para full & -& 93.05$^{\dagger}$& -& 89.04$^{\dagger}$& -& 84.86$^{\dagger}$& -& 79.54\\
            w/ sent-link full & -& 93.14$^{\dagger}$& -& 89.10$^{\dagger}$& -& 84.94$^{\dagger}$& -& 79.64$^{\dagger}$\\
            w/ sent full & -& 93.02$^{\dagger}$& -& 89.17$^{\dagger}$& -& 84.88$^{\dagger}$& -& 79.70$^{\dagger}$\\
            w/ para KL & 93.09$^{\dagger}$& 93.15$^{\dagger}$& 89.05$^{\dagger}$& 89.39$^{\dagger}$& 84.90$^{\dagger}$& 85.36$^{\dagger}$& 79.91$^{\dagger}$& 80.27$^{\dagger}$\\
            w/ sent-link KL & 93.16$^{\dagger}$& 93.17$^{\dagger}$& 89.09$^{\dagger}$& 89.48$^{\dagger}$& 85.27$^{\dagger}$& 85.53$^{\dagger}$& 79.92$^{\dagger}$& 80.30$^{\dagger}$\\
            w/ sent KL & 93.12$^{\dagger}$& 93.08$^{\dagger}$& 89.21$^{\dagger}$& 89.40$^{\dagger}$& 84.94$^{\dagger}$& 85.32$^{\dagger}$& 79.95$^{\dagger}$& 80.41$^{\dagger}$\\
            w/ para L2 & 93.08$^{\dagger}$& 93.12$^{\dagger}$& 89.08$^{\dagger}$& 89.40$^{\dagger}$& 85.01$^{\dagger}$& 85.35$^{\dagger}$& 79.88$^{\dagger}$& 80.33$^{\dagger}$\\
            w/ sent-link L2 & 93.07$^{\dagger}$& 93.26$^{\dagger}$& 89.16$^{\dagger}$& 89.49$^{\dagger}$& 85.04$^{\dagger}$& 85.48$^{\dagger}$& 79.96$^{\dagger}$& 80.28$^{\dagger}$\\
            w/ sent L2 & 93.07$^{\dagger}$& 93.19$^{\dagger}$& 89.14$^{\dagger}$& 89.51$^{\dagger}$& 85.13$^{\dagger}$& 85.33$^{\dagger}$& 79.95$^{\dagger}$& 80.30$^{\dagger}$\\
            \hline
        \end{tabular} }
    }
    \begin{threeparttable}
        \begin{tablenotes}
        \footnotesize
         \item[] The $^{\dagger}$ symbol indicates that the corresponding model demonstrates a significant improvement over the baseline model (\emph{noisy text only} w/o context) at a confidence level of $p<0.01$.
        \end{tablenotes}
    \end{threeparttable}
\end{table*}

\begin{table*}[tb]
    \setlength{\tabcolsep}{1.5pt}
    \centering
    \caption{NER performance with OCR noise}
    \label{the NER performance of OCR noise}
    \subcaptionbox{WNUT-17}
    {
        \resizebox{0.483\linewidth}{!}{
        \begin{tabular}{c|cc|cc|cc|cc|cc}
            \hline
            \rule{0pt}{12pt}
            \multirow{2}*{OCR type} &
            \multicolumn{2}{c|}{no} & \multicolumn{2}{c|}{OCR\_1} & \multicolumn{2}{c|}{OCR\_2} & \multicolumn{2}{c|}{OCR\_3}& \multicolumn{2}{c}{OCR\_4}\\
            \cline{2-11}
            \rule{0pt}{12pt}
            & OV & RV & OV & RV & OV & RV & OV & RV & OV & RV \\
            \hline
            \multicolumn{1}{l|}{\emph{Baselines: }} & & & & & & & & & & \\
            \multicolumn{1}{l|}{\emph{noisy text only}} & & & & & & & & & & \\
            w/o context & 58.21& -& 53.95& -& 50.00& -& 45.63& -&35.24& -\\
            \hline
            \multicolumn{1}{l|}{\emph{correct}} & & & & & & & & & & \\
            \multicolumn{1}{l|}{\emph{w/o context }} & & & & & & & & & & \\
            natas & -& -& 45.27& -& 42.06& -& 38.76& -& 28.05& -\\
            hunspell & -& -& 50.66& -& 45.60& -& 40.44& -& 31.90& -\\
            \hline
            \multicolumn{1}{l|}{\emph{with gold text}} & & & & & & & & \\
            w/ gold full & -& -& -& 53.99& -& 52.37$^{\dagger}$& -& 52.07$^{\dagger}$& -& 48.59$^{\dagger}$\\
            w/ gold KL & -& -& 55.25$^{\dagger}$& 55.13$^{\dagger}$& 50.83$^{\dagger}$& 53.06$^{\dagger}$& 45.86$^{\dagger}$& 50.98$^{\dagger}$& 38.31$^{\dagger}$& 48.44$^{\dagger}$\\
            w/ gold L2 & -& -& 54.61$^{\dagger}$& 54.55$^{\dagger}$& 51.18$^{\dagger}$& 53.72$^{\dagger}$& 45.89$^{\dagger}$& 51.43$^{\dagger}$& 37.95$^{\dagger}$& 48.27$^{\dagger}$\\
            \hline
            \multicolumn{1}{l|}{\emph{self-retrieval}} & & & & & & & & \\
            w/ self full & -& 58.90$^{\dagger}$& -& 54.72$^{\dagger}$& -& 51.30$^{\dagger}$& -& 45.68$^{\dagger}$& -& 37.18$^{\dagger}$\\
            w/ self KL & 59.59$^{\dagger}$& 59.87$^{\dagger}$& 55.07$^{\dagger}$& 55.11$^{\dagger}$& 51.13$^{\dagger}$& 51.21$^{\dagger}$& 45.92$^{\dagger}$& 45.99$^{\dagger}$& 37.47$^{\dagger}$& 37.98$^{\dagger}$\\
            w/ self L2 & 59.02$^{\dagger}$& 59.21$^{\dagger}$& 55.21$^{\dagger}$& 54.73$^{\dagger}$& 51.04$^{\dagger}$& 51.52$^{\dagger}$& 45.98$^{\dagger}$& 46.09$^{\dagger}$& 37.59$^{\dagger}$& 37.42$^{\dagger}$\\
            \hline
            \multicolumn{1}{l|}{\emph{BM25-retrieval}} & & & & & & & & \\
            w/ para full & -& 59.29$^{\dagger}$& -& 54.53$^{\dagger}$& -& 51.38$^{\dagger}$& -& 46.21$^{\dagger}$& -& 38.01$^{\dagger}$\\
            w/ sent-link full & -& 59.88$^{\dagger}$& -& 56.33$^{\dagger}$& -& 51.03$^{\dagger}$& -& 46.30$^{\dagger}$& -& 37.96$^{\dagger}$\\
            w/ sent full & -& 59.15$^{\dagger}$& -& 55.26$^{\dagger}$& -& 51.53$^{\dagger}$& -& 46.10$^{\dagger}$& -& 37.75$^{\dagger}$\\
            w/ para KL & 59.39$^{\dagger}$& 60.02$^{\dagger}$& 55.05$^{\dagger}$& 55.38$^{\dagger}$& 51.13$^{\dagger}$& 52.28$^{\dagger}$& 46.36$^{\dagger}$& 46.37$^{\dagger}$& 37.76$^{\dagger}$& 38.49$^{\dagger}$\\
            w/ sent-link KL & 59.54$^{\dagger}$& 60.61$^{\dagger}$& 54.90$^{\dagger}$& 55.76$^{\dagger}$& 50.99$^{\dagger}$& 52.01$^{\dagger}$& 45.93$^{\dagger}$& 45.94$^{\dagger}$& 37.19$^{\dagger}$& 38.23$^{\dagger}$\\
            w/ sent KL & 59.21$^{\dagger}$& 59.65$^{\dagger}$& 54.95$^{\dagger}$& 55.35$^{\dagger}$& 51.00$^{\dagger}$& 51.58$^{\dagger}$& 45.81$^{\dagger}$& 46.80$^{\dagger}$& 37.85$^{\dagger}$& 38.25$^{\dagger}$\\
            w/ para L2 & 59.94$^{\dagger}$& 59.92$^{\dagger}$& 54.99$^{\dagger}$& 55.16$^{\dagger}$& 51.10$^{\dagger}$& 52.09$^{\dagger}$& 46.51$^{\dagger}$& 46.54$^{\dagger}$& 37.33$^{\dagger}$& 37.70$^{\dagger}$\\
            w/ sent-link L2 & 59.02$^{\dagger}$& 59.05$^{\dagger}$& 55.06$^{\dagger}$& 56.10$^{\dagger}$& 51.47$^{\dagger}$& 52.54$^{\dagger}$& 46.86$^{\dagger}$& 47.30$^{\dagger}$& 37.31$^{\dagger}$& 37.75$^{\dagger}$\\
            w/ sent L2 & 59.10$^{\dagger}$& 59.39$^{\dagger}$& 55.05$^{\dagger}$& 55.51$^{\dagger}$& 51.18$^{\dagger}$& 52.73$^{\dagger}$& 46.56$^{\dagger}$& 47.25$^{\dagger}$& 37.36$^{\dagger}$& 37.39$^{\dagger}$\\
            \hline
            \multicolumn{1}{l|}{\emph{dense-retrieval}} & & & & & & & & \\
            w/ para full & -& 59.09$^{\dagger}$& -& 54.73$^{\dagger}$& -& 51.70$^{\dagger}$& -& 47.18$^{\dagger}$& -& 37.33$^{\dagger}$\\
            w/ sent-link full & -& 58.95$^{\dagger}$& -& 54.42$^{\dagger}$& -& 51.06$^{\dagger}$& -& 47.77$^{\dagger}$& -& 37.30$^{\dagger}$\\
            w/ sent full & -& 59.23$^{\dagger}$& -& 54.38$^{\dagger}$& -& 51.49$^{\dagger}$& -& 47.10$^{\dagger}$& -& 37.92$^{\dagger}$\\
            w/ para KL & 59.21$^{\dagger}$& 59.70$^{\dagger}$& 55.24$^{\dagger}$& 54.96$^{\dagger}$& 51.46$^{\dagger}$& 51.28$^{\dagger}$& 46.72$^{\dagger}$& 46.79$^{\dagger}$& 37.99$^{\dagger}$& 38.11$^{\dagger}$\\
            w/ sent-link KL & 59.99$^{\dagger}$& 60.04$^{\dagger}$& 54.89$^{\dagger}$& 55.11$^{\dagger}$& 51.13$^{\dagger}$& 51.50$^{\dagger}$& 46.68$^{\dagger}$& 47.63$^{\dagger}$& 37.31$^{\dagger}$& 37.72$^{\dagger}$\\
            w/ sent KL & 59.26$^{\dagger}$& 59.79$^{\dagger}$& 54.54$^{\dagger}$& 53.91$^{\dagger}$& 51.70$^{\dagger}$& 51.57$^{\dagger}$& 46.19$^{\dagger}$& 46.87$^{\dagger}$& 37.42$^{\dagger}$& 37.20$^{\dagger}$\\
            w/ para L2 & 59.57$^{\dagger}$& 59.46$^{\dagger}$& 55.70$^{\dagger}$& 55.28$^{\dagger}$& 51.57$^{\dagger}$& 51.16$^{\dagger}$& 46.92$^{\dagger}$& 46.46$^{\dagger}$& 38.35$^{\dagger}$& 37.70$^{\dagger}$\\
            w/ sent-link L2 & 58.98$^{\dagger}$& 59.29$^{\dagger}$& 55.39$^{\dagger}$& 55.07$^{\dagger}$& 50.81$^{\dagger}$& 51.28$^{\dagger}$& 46.23$^{\dagger}$& 46.62$^{\dagger}$& 37.22$^{\dagger}$& 37.07$^{\dagger}$\\
            w/ sent L2 & 59.04$^{\dagger}$& 58.97$^{\dagger}$& 54.79$^{\dagger}$& 54.09$^{\dagger}$& 50.76$^{\dagger}$& 51.34$^{\dagger}$& 46.55$^{\dagger}$& 46.96$^{\dagger}$& 37.30$^{\dagger}$& 37.62$^{\dagger}$\\
            \hline
        \end{tabular} }
    }
    \hfill
    \subcaptionbox{CoNLL-03}
    {
        \resizebox{0.483\linewidth}{!}{
        \begin{tabular}{c|cc|cc|cc|cc|cc}
            \hline
            \rule{0pt}{12pt}
            \multirow{2}*{OCR type} &
            \multicolumn{2}{c|}{no} & \multicolumn{2}{c|}{OCR\_1} & \multicolumn{2}{c|}{OCR\_2} & \multicolumn{2}{c|}{OCR\_3}& \multicolumn{2}{c}{OCR\_4}\\
            \cline{2-11}
            \rule{0pt}{12pt}
            & OV & RV & OV & RV & OV & RV & OV & RV & OV & RV \\
            \hline
            \multicolumn{1}{l|}{\emph{Baselines: }} & & & & & & & & \\
            \multicolumn{1}{l|}{\emph{noisy text only}} & & & & & & & & \\
            w/o context & 92.98& -& 92.88& -& 91.45& -& 86.77& -& 79.83& -\\
            \hline
            \multicolumn{1}{l|}{\emph{correct}} & & & & & & & & & & \\
            \multicolumn{1}{l|}{\emph{w/o context }} & & & & & & & & & & \\
            natas & -& -& 89.22& -& 87.04& -& 82.78& -& 75.19& -\\
            hunspell & -& -& 91.22& -& 89.71& -& 84.91& -& 77.55& -\\
            \hline
            \multicolumn{1}{l|}{\emph{with gold text}} & & & & & & & & \\
            w/ gold full & -& -& -& 93.06$^{\dagger}$& -& 92.90$^{\dagger}$& -& 91.60$^{\dagger}$& -& 89.53$^{\dagger}$\\
            w/ gold KL & -& -& 92.94$^{\dagger}$& 93.00$^{\dagger}$& 91.51$^{\dagger}$& 92.79$^{\dagger}$& 87.01$^{\dagger}$& 91.58$^{\dagger}$& 80.09$^{\dagger}$& 89.34$^{\dagger}$\\
            w/ gold L2 & -& -& 92.91$^{\dagger}$& 92.97$^{\dagger}$& 91.52$^{\dagger}$& 92.85$^{\dagger}$& 87.24$^{\dagger}$& 91.68$^{\dagger}$& 80.13$^{\dagger}$& 89.30$^{\dagger}$\\
            \hline
            \multicolumn{1}{l|}{\emph{self-retrieval}} & & & & & & & & \\
            w/ self full & -& 93.10$^{\dagger}$& -& 92.99$^{\dagger}$& -& 91.48$^{\dagger}$& -& 86.91$^{\dagger}$& -& 80.37$^{\dagger}$\\
            w/ self KL & 93.21$^{\dagger}$& 93.46$^{\dagger}$& 92.99$^{\dagger}$& 93.27$^{\dagger}$& 91.48$^{\dagger}$& 91.76$^{\dagger}$& 86.87$^{\dagger}$& 87.33$^{\dagger}$& 79.98$^{\dagger}$& 80.79$^{\dagger}$\\
            w/ self L2 & 93.13$^{\dagger}$& 93.54$^{\dagger}$& 92.96$^{\dagger}$& 93.19$^{\dagger}$& 91.47$^{\dagger}$& 91.81$^{\dagger}$& 86.84$^{\dagger}$& 87.32$^{\dagger}$& 79.86$^{\dagger}$& 80.72$^{\dagger}$\\
            \hline
            \multicolumn{1}{l|}{\emph{BM25-retrieval}} & & & & & & & & \\
            w/ para full & -& 93.33$^{\dagger}$& -& 93.21$^{\dagger}$& -& 91.78$^{\dagger}$& -& 87.18$^{\dagger}$& -& 80.15$^{\dagger}$\\
            w/ sent-link full & -& 93.27$^{\dagger}$& -& 93.23$^{\dagger}$& -& 91.79$^{\dagger}$& -& 87.10$^{\dagger}$& -& 80.24$^{\dagger}$\\
            w/ sent full & -& 93.33$^{\dagger}$& -& 93.08$^{\dagger}$& -& 91.63$^{\dagger}$& -& 87.18$^{\dagger}$& -& 80.27$^{\dagger}$\\
            w/ para KL &93.07$^{\dagger}$ &93.26$^{\dagger}$ &92.91$^{\dagger}$ &93.00$^{\dagger}$ &91.47$^{\dagger}$ &91.71$^{\dagger}$ &87.05$^{\dagger}$ &87.34$^{\dagger}$ &79.77 &80.25$^{\dagger}$\\
            w/ sent-link KL &93.08$^{\dagger}$ &93.34$^{\dagger}$ &93.10$^{\dagger}$ &93.47$^{\dagger}$ &91.47$^{\dagger}$ &91.90$^{\dagger}$ &86.93$^{\dagger}$ &87.34$^{\dagger}$ &79.39 &80.06$^{\dagger}$\\
            w/ sent KL &93.05$^{\dagger}$ &93.48$^{\dagger}$ &93.01$^{\dagger}$ &93.24$^{\dagger}$ &91.57$^{\dagger}$ &91.96$^{\dagger}$ &87.06$^{\dagger}$ &87.31$^{\dagger}$ &79.26 &79.87$^{\dagger}$\\
            w/ para L2 &93.10$^{\dagger}$ &93.31$^{\dagger}$ &93.01$^{\dagger}$ &93.23$^{\dagger}$ &91.66$^{\dagger}$ &91.85$^{\dagger}$ &86.95$^{\dagger}$ &87.18$^{\dagger}$ &79.70 &80.22$^{\dagger}$\\
            w/ sent-link L2 &93.14$^{\dagger}$ &93.53$^{\dagger}$ &92.90$^{\dagger}$ &93.30$^{\dagger}$ &91.51$^{\dagger}$ &91.74$^{\dagger}$ &86.88$^{\dagger}$ &87.36$^{\dagger}$ &79.60 &80.64$^{\dagger}$\\
            w/ sent L2 &93.19$^{\dagger}$ &93.42$^{\dagger}$ &92.94$^{\dagger}$ &93.07$^{\dagger}$ &91.53$^{\dagger}$ &91.93$^{\dagger}$ &87.06$^{\dagger}$ &87.44$^{\dagger}$ &79.76 &80.49$^{\dagger}$\\
            \hline
            \multicolumn{1}{l|}{\emph{dense-retrieval}} & & & & & & & & & & \\
            w/ para full & -& 93.33$^{\dagger}$& -& 93.21$^{\dagger}$& -& 91.68$^{\dagger}$& -& 87.18$^{\dagger}$& -& 80.34$^{\dagger}$\\
            w/ sent-link full & -& 93.27$^{\dagger}$& -& 93.23$^{\dagger}$& -& 91.86$^{\dagger}$& -& 87.14$^{\dagger}$& -& 80.06$^{\dagger}$\\
            w/ sent full & -& 93.35$^{\dagger}$& -& 93.01$^{\dagger}$& -& 91.50$^{\dagger}$& -& 87.21$^{\dagger}$& -& 80.37$^{\dagger}$\\
            w/ para KL &93.26$^{\dagger}$ &93.27$^{\dagger}$ &92.98$^{\dagger}$ &92.92$^{\dagger}$ &91.62$^{\dagger}$ &91.83$^{\dagger}$ &87.25$^{\dagger}$ &87.44$^{\dagger}$ &80.07$^{\dagger}$ &80.21$^{\dagger}$\\
            w/ sent-link KL &93.21$^{\dagger}$ &93.43$^{\dagger}$ &93.04$^{\dagger}$ &93.26$^{\dagger}$ &91.51$^{\dagger}$ &91.87$^{\dagger}$ &87.18$^{\dagger}$ &87.36$^{\dagger}$ &79.86$^{\dagger}$ &80.19$^{\dagger}$\\
            w/ sent KL &93.08$^{\dagger}$ &93.36$^{\dagger}$ &93.06$^{\dagger}$ &93.13$^{\dagger}$ &91.56$^{\dagger}$ &91.83$^{\dagger}$ &87.11$^{\dagger}$ &87.30$^{\dagger}$ &79.98$^{\dagger}$ &80.29$^{\dagger}$\\
            w/ para L2 &93.09$^{\dagger}$ &93.23$^{\dagger}$ &92.89$^{\dagger}$ &92.97$^{\dagger}$ &91.76$^{\dagger}$ &91.82$^{\dagger}$ &87.04$^{\dagger}$ &87.37$^{\dagger}$ &79.87$^{\dagger}$ &80.10$^{\dagger}$\\
            w/ sent-link L2 &93.14$^{\dagger}$ &93.53$^{\dagger}$ &92.90$^{\dagger}$ &93.28$^{\dagger}$ &91.57$^{\dagger}$ &91.85$^{\dagger}$ &87.03$^{\dagger}$ &87.41$^{\dagger}$ &79.86$^{\dagger}$ &80.06$^{\dagger}$\\
            w/ sent L2 &93.07$^{\dagger}$ &93.23$^{\dagger}$ &93.05$^{\dagger}$ &93.10$^{\dagger}$ &91.59$^{\dagger}$ &91.93$^{\dagger}$ &87.08$^{\dagger}$ &87.26$^{\dagger}$ &80.00$^{\dagger}$ &80.44$^{\dagger}$\\
            \hline
        \end{tabular} }
    }
    \begin{threeparttable}
        \begin{tablenotes}
        \footnotesize
         \item[] The $^{\dagger}$ symbol indicates that the corresponding model demonstrates a significant improvement over the baseline model (\emph{noisy text only} w/o context) at a confidence level of $p<0.01$.
        \end{tablenotes}
    \end{threeparttable}
\end{table*}

\subsection{Retrieval Module Configuration}

\subsubsection{BM25 retrieval (sparse retrieval)}
We employ the 2022-07-01 version of the Wikipedia data dump\footnote{https://dumps.wikipedia.org/} as our primary data source, which is subsequently converted into plain text format for efficient processing. For the implementation of the BM25 retrieval algorithm, we utilize Elasticsearch\footnote{https://www.elastic.co} to index and retrieve sentences within the Wikipedia corpus. In this configuration, the smallest searchable unit is defined as a single sentence. Furthermore, the matching algorithm is configured to ``match'', signifying that the retrieved results must contain a minimum of one word from the query sentence.

\subsubsection{Dense retrieval}
In our dense retrieval experiment, we utilize the entire Wikipedia corpus to retrieve, consistent with the BM25 retrieval method, and an encoder trained using the unsupervised SimCSE approach \citep{gao-etal-2021-simcse}. The encoder is fine-tuned on the XLM-RoBERTa-large model through contrastive learning, employing simulated noisy text and clean text from Wikipedia. Two distinct noise induction techniques, spelling mistakes and OCR errors, are applied separately. In order to address challenges over RAM usage and computational speed and inspired by BERT-whitening \citep{su2021whitening}, we generate sentence embeddings, perform PCA for dimensionality reduction, and save the embeddings as npy files. We leverage the FAISS library\footnote{https://github.com/facebookresearch/faiss} to implement the Inverted File System \citep{mazur1979inverted}  using the index\_factory function, incorporating the Hierarchical Navigable Small World (HNSW) technique \citep{malkov2018efficient} and the cosine distance metric. Further details can be found in the Appendix~\ref{Appendix the details of dense retrieval}.

\subsubsection{BERTScore retrieval (self retrieval)}
The official implementation of BERTScore \footnote{https://github.com/Tiiiger/bert\_score} can only calculate paired inputs, 
which is very slow for a large number of query and target sentences. We modify the source code by first encoding all the query and target sentences to contextual token embeddings, then doing matching for each pair of sentences, and finally outputting their similarity scores.
It is worth noting that each noisy text is tested individually and does not interfere with each other, i.e., inductive inference. Therefore, we independently retrieve texts for the training, development, and testing sets from the same source corpus, i.e., the training set.

\subsubsection{Postprocessing}
We use top-10 retrieved results. For better utilization of retrieval results of Wikipedia, we follow previous work \citep{wang-etal-2022-damo} and use three options, i.e., using the matched paragraph, using the matched sentence, and using the matched sentence but removing wiki anchors, i.e., mention hyperlinks, which can provide rich and useful clues for mention detection in NER. If there is a hyperlink $[\emph{A}\rightarrow \textup{B}]$, we will transfer word $\emph{A}$ in the sentence to $<e:\textup{B}>\emph{A}</e>$. For each retrieval result from Wikipedia, we concatenate it with the corresponding title to form the final retrieval result. To get the matched paragraph for dense retrieval, we can use Elasticsearch by the matching algorithm ``term'', i.e., complete matching, to search Wikipedia. For the results retrieved from the noisy dataset itself, we do not do post-processing.

\subsection{NER Module Configuration}
We use XLM-RoBERTa-large \citep{conneau-etal-2020-unsupervised} as the pre-trained transformer-based model, tune the learning rate of the transformer within $[1\times10^{6}, 1\times10^{5}]$, and use grid search to find the ratio of CRF learning rate and transformer learning rate within $[3000, 12000]$.

\subsection{Main Results}
We employ entity-level F1 scores as the evaluation metric for Named Entity Recognition (NER). For each experimental setting, we run four different seeds and then average the results. The NER performance is shown in Table~\ref{the NER performance of spelling mistakes} and Table~\ref{the NER performance of OCR noise} for spelling mistakes and OCR noise respectively.
For the first row, $0$ or $no$ denotes the original clean text without noise and the other denote different types of noisy text; $OV$ is the original noisy text view and $RV$ is the retrieval-based view.
There are many settings that correspond to each column, as detailed below:
\begin{itemize}

\item $noisy\ text\ only\ w/o\ context$ only uses the original noisy text without the retrieved text.

\item $w/o\ context\ natas/hunspell$ only uses the original noisy text without the retrieved text but the noisy text is corrected before inputting by natas/hunspell.

\item $w/o\ context\ natas/hunspell+check$ adds a check process which only retains words corrected rightly but keeps the original noisy words if corrected wrongly.

\item $w/\ gold/self$ uses both the original noisy text and the external context which is the corresponding gold text or the self retrieval context.

\item $w/\ para/sent/sent-link$ uses both the original noisy text and the external context which is the matched paragraph, the matched sentence and the matched sentence without wiki anchors.

\item $full$ denotes using only the original noisy text concatenated with retrieved context.

\item $KL$ denotes using multi-view learning with the KL loss.

\item $L2$ denotes using multi-view learning with the L2 loss.

\end{itemize}

The baseline model is $noisy\ text\ only\ w/o\ context$. We are the first approach to robust NER with only noisy training text and without its corresponding gold text, so there is not SoTA.  We employ the Deep Dominance method \citep{dror-etal-2019-deep} to perform statistical significance testing. In Tables~\ref{the NER performance of spelling mistakes} and \ref{the NER performance of OCR noise}, it is evident that our proposed method exhibits a significant improvement in NER performance over the baseline.

We also use the post-correction model NATAS \footnote{\url{https://github.com/mikahama/natas}} \citep{hamalainen-hengchen-2019-paft} and the spell checker Hunspell \footnote{\url{http://hunspell.github.io/}} to correct the words before feeding them into the NER model, but find that the performance is worse than only using noisy text, which is probably because of erroneous corrections. Furthermore, when we keep the corrected words which are error-corrected the same as the gold words while not correcting them if error-correcting wrongly, which can be seen in the setting of $w/o\ context\ natas/hunspell+check$, then the performance is about the same as the original noisy text. This suggests that erroneous correction of words may result in a detrimental effect on the NER performance due to the self-attention mechanism inherent in transformer-based models by the words of erroneous correction. More statistics of the correction experiment can be seen in Appendix~\ref{Appendix the statistics of errors in the correction experiment}.

From the results in the setting of $with\ gold\ text$, we can see that if we concatenate noisy text with the corresponding gold text, then the performance is competitive with inputting clean text without noise, which shows that the self-attention mechanism of transformer-based model can improve the representation of noisy text.

The reason why the original WNUT-17 dataset has worse NER performance compared with the original CoNLL-03 dataset is due to the former being sourced from Twitter, which inherently has more noise, while the latter is sourced from news articles, which are cleaner. Hence, when using retrieval augmentation, the improvement of NER performance in WNUT-17 is more significant.

The performance of the three retrieval approaches demonstrates no discernible uniformity and not regular, as their outcomes exhibit considerable diversity across disparate data domains, unique noise introduction techniques, and fluctuating noise intensities.

\subsection{Comparing the model using noisy text vs. using clean text}
We do a cross-experiment using noisy text and clean text. The results can be seen in Table~\ref{cross-experience between spelling mistakes text and clean text} and Table~\ref{cross-experience between OCR erroneous text and clean text} for spelling mistakes and OCR noise respectively. The model trained on noisy text can perform well on clean text, which indicates that the robust NER model is capable of
recognizing both noisy text and clean text.

\begin{table}[tb]
    \centering
    \caption{Cross-experiment on spelling mistakes text and clean text}
    \label{cross-experience between spelling mistakes text and clean text}
        \resizebox{0.95\linewidth}{!}{
        \begin{tabular}{c|cccc}
            \hline
            \rule{0pt}{12pt}
            \diagbox{dataset}{model}&clean&typos\_0.1&typos\_0.2&typos\_0.3 \\
            \hline
            \\[-6pt]
            \multicolumn{1}{l|}{\emph{WNUT-17}} & & & & \\
            clean & 58.21& 56.00& 50.15& 34.06\\
            typos\_0.1 & 44.0& 46.26& & \\
            typos\_0.2 & 26.29& & 33.45& \\
            typos\_0.3 & 12.48& & & 24.72\\
            \hline
            \\[-6pt]
            \multicolumn{1}{l|}{\emph{CoNLL-03}} & & & & \\
            clean &92.98& 92.73& 92.42& 92.00 \\
            typos\_0.1 &80.72& 88.82& &  \\
            typos\_0.2 &68.27& & 84.37&  \\
            typos\_0.3 &54.02& & & 79.53 \\
            \hline
        \end{tabular} }
\end{table}

\begin{table}[tb]
    \centering
    \caption{Cross-experiment on OCR erroneous text and clean text}
    \label{cross-experience between OCR erroneous text and clean text}
        \resizebox{0.95\linewidth}{!}{
        \begin{tabular}{c|ccccc}
            \hline
            \rule{0pt}{12pt}
            \diagbox{dataset}{model}&clean&OCR\_1&OCR\_2&OCR\_3&OCR\_4 \\
            \hline
            \\[-6pt]
            \multicolumn{1}{l|}{\emph{WNUT-17}} & & & & & \\
            clean & 58.21& 54.95& 54.44& 54.92& 51.82\\
            OCR\_1 & 56.79& 53.95& & & \\
            OCR\_2 & 51.83& & 50.00& & \\
            OCR\_3 & 47.74& & & 45.63& \\
            OCR\_4 & 37.95& & & & 35.24\\
            \hline
            \\[-6pt]
            \multicolumn{1}{l|}{\emph{CoNLL-03}} & & & & & \\
            clean & 92.98& 93.13& 93.12& 92.80& 92.65\\
            OCR1 & 91.86& 92.88& & & \\
            OCR2 & 88.17& & 91.45& & \\
            OCR3 & 81.78& & & 86.77& \\
            OCR4 & 71.46& & & & 79.83\\
            \hline
        \end{tabular} }
\end{table}

\section{Related Work}
\paragraph{Robust representations}
One way to make a model more robust is to make the representations less sensitive to noisy input. \citep{zheng2016improving, cheng-etal-2018-towards} add noise to input in each epoch and use pairs of noisy input and the corresponding gold input for training. \citep{piktus-etal-2019-misspelling, jones-etal-2020-robust} improve robust encoding by embedding misspellings close to their correct variants and mapping sentences to a smaller discrete space. However, these methods all require gold text, which is impractical in real scenarios.

\paragraph{Noisy sequence labeling}
There are a few previous studies of noisy sequence labeling \citep{namysl-etal-2020-nat, namysl-etal-2021-empirical}, which use gold text to improve the performance on noisy text. However, gold text may be unavailable or undesirable in many scenarios, such as there are only noisy text in the real scenario. In contrast, our work does not rely on gold text. In the OCR errors setting, \citep{namysl-etal-2021-empirical} make the length of noisy text and that of the corresponding gold text the same, which is unrealistic, while in our setting the number of labels is the same as the number of recognized words after OCR.

\paragraph{Error correction}
A straightforward method to handle noisy text is to correct it. There are many studies on error correction, such as spelling-correction \citep{etoori-etal-2018-automatic}, post-OCR correction \citep{hamalainen-hengchen-2019-paft}, and sequence-to-sequence model \citep{ge2018reaching}. However, these error correction models rely on priori error correction knowledge, which may be different from the given datasets, so the error correction models do not work well in our experiments.

\paragraph{Retrieval augmentation}
Many previous studies use retrieval augmentation to improve specific task performance, such as language generation \citep{lewis2020retrieval} and relation extraction \citep{zhang-2021-readsre}. For the NER task, \citep{wang-etal-2021-improving, wang-etal-2022-ita, Wang2022NamedEA} use external retrieved context to improve the performance of NER and multi-modal NER respectively.

\section{Conclusion}
In this paper, we propose a robust Named Entity Recognition (NER) model that is able to effectively handle noisy text without the need for corresponding gold text during training. Our approach utilizes retrieval augmentation and we jointly train a retrieval-based view and an original noisy text view using multi-view learning. Our approach results in significant improvements in handling both misspelling mistakes and OCR errors.



\appendix

\section{the details of dense retrieval}
\label{Appendix the details of dense retrieval}
The dense retrieval approach utilizes the entire Wikipedia, same as BM25 retrieval. Our encoder is trained on a dataset of randomly sampled sentences from Wikipedia, preprocessed using the unsupervised SimCSE method \citep{gao-etal-2021-simcse}, with a sentence size set to $10^{6}$. To simulate noisy text, we introduce noise into these sentences and partition the dataset into train, development, and test sets in an 8:1:1 ratio. Our experiments involve two distinct noise induction methods, namely spelling mistakes and OCR errors, conducted separately.

For spelling mistakes, we introduce noise at random levels ranging from 0 to 0.4. In the case of OCR errors, we incorporate random fronts and random OCR error induction approaches. We employ the XLM-RoBERTa-large model, fine-tuned using contrastive learning on the simulated noisy text and the corresponding original clean text sourced from Wikipedia. The sentence embeddings are generated by averaging the embeddings of the first and last layers. Model selection is based on the $(Recall@1+Recall@4+Recall@16+Recall@64)$ metric, which measures the retrieval performance of the corresponding gold text in Wikipedia using the simulated noisy text in the development set. The contrastive learning model is trained with a temperature hyperparameter of 0.3 and an epoch of 1. Detailed training information can be found in Appendix~\ref{Appendix the details of contrastive learning training}.

Upon completion of encoder training, we utilize dense retrieval to search the entire Wikipedia using the FAISS library. As Wikipedia comprises approximately $2.29\times10^{8}$ sentences, dense retrieval poses significant challenges in terms of RAM usage and computational speed. To address these issues, we encode the Wikipedia sentences into 1024-dimensional embeddings and store them as npy files on disk, which has greater capacity than RAM. We then perform dense retrieval on these files. Drawing inspiration from BERT-whitening \citep{su2021whitening}, we employ PCA to reduce the embedding dimensions from 1024 to 128, enabling the dense retrieval model to operate within a 250 GB RAM constraint. The PCA model is trained on a random sample of 20\% of Wikipedia sentences, approximately $4.57\times10^{7}$ in number.

To expedite the retrieval process, we implement the Inverted File System (IVF) \citep{mazur1979inverted} using the index\_factory function in FAISS. Specifically, we configure the index\_factory(1024, ``PCA128, IVF262144\_HNSW32, Flat'') to represent 262,144 clustering centers. The base index relies on the Hierarchical Navigable Small World (HNSW) \citep{malkov2018efficient} graph-based approach, utilizing 32 neighbors. As HNSW cannot be accelerated by GPU, we train on GPU and execute other operations, such as retrieval, on CPU. Our model leverages pre-normalized embeddings and the L2 distance measure to evaluate embedding similarity, which is widely used and well-supported in FAISS, rendering it equivalent to cosine distance. We set the nprobe hyperparameter to 64, allowing the model to perform fine-grained retrieval in the 64 nearest clusters following the $knn$ algorithm applied to the clustering centers. The nprobe value balances recall and search efficiency.

\section{the details of contrastive learning training}
\label{Appendix the details of contrastive learning training}
The details of contrastive learning training is as Table~\ref{contrastive learning training in spelling mistakes datasets} and Table~\ref{contrastive learning training in OCR errors datasets}. $not\ train$ is the XLM-RoBERTa-large model which is not fined tune. $recall\_avg=(recall@1+recall@4+recall@16+recall@64)/4$, $typos\_mix$ or $OCR\_mix$ is the uniform mixing noisy dataset. Each column in the experimental results represents a distinct setting, denoting the effectiveness of the model trained through contrastive learning in retrieving various forms of noisy or mixed noisy text, as measured by the $recall@k$ metric.

\begin{table}[tb]
    \centering
    \caption{details of contrastive learning training in spelling mistakes datasets}
    \label{contrastive learning training in spelling mistakes datasets}
        \resizebox{0.95\linewidth}{!}{
        \begin{tabular}{c|ccc|ccc|c}
        \hline
        \multicolumn{1}{c|}{batchsize}& \multicolumn{3}{c|}{64} & \multicolumn{3}{c|}{512} & \multirow{2}{*}{not train}   \\
        learning rate & 1e-5 & 3e-5 & 5e-5     & 1e-5 & 3e-5 & 5e-5      &                             \\
        \hline
        \multicolumn{1}{l|}{\emph{typos\_0.1}} & & & & & & & \\
        recall@1      &0.9943&0.9279&0.9240&0.9924&0.9932&0.9939&0.4657\\
        recall@4      &0.9961&0.9382&0.9345&0.9945&0.9953&0.9960&0.5010\\
        recall@16     &0.9966&0.9453&0.9412&0.9951&0.9960&0.9965&0.5287\\
        recall@64     &0.9968&0.9512&0.9470&0.9953&0.9961&0.9967&0.5547\\
        recall\_avg   &0.9959&0.9407&0.9367&0.9943&0.9952&0.9958&0.5125\\
        \hline
        \multicolumn{1}{l|}{\emph{typos\_0.2}} & & & & & & & \\
        recall@1      &0.9799&0.8014&0.8047&0.9701&0.9765&0.9795&0.1250\\
        recall@4      &0.9848&0.8188&0.8215&0.9768&0.9822&0.9848&0.1545\\
        recall@16     &0.9862&0.8325&0.8321&0.9788&0.9837&0.9862&0.1854\\
        recall@64     &0.9867&0.8474&0.8424&0.9796&0.9842&0.9866&0.2219\\
        recall\_avg   &0.9844&0.8250&0.8252&0.9763&0.9816&0.9843&0.1717\\
        \hline
        \multicolumn{1}{l|}{\emph{typos\_0.3}} & & & & & & & \\
        recall@1      &0.9496&0.7291&0.7296&0.9199&0.9411&0.9483&0.0444\\
        recall@4      &0.9623&0.7482&0.7555&0.9415&0.9561&0.9626&0.0583\\
        recall@16     &0.9659&0.7605&0.7685&0.9488&0.9606&0.9668&0.0763\\
        recall@64     &0.9674&0.7730&0.7787&0.9519&0.9623&0.9682&0.1008\\
        recall\_avg   &0.9613&0.7527&0.7581&0.9405&0.9550&0.9615&0.0700\\
        \hline
        \multicolumn{1}{l|}{\emph{typos\_mix}} & & & & & & & \\
        recall@1      &0.9676&0.8233&0.8220&0.9500&0.9626&0.9666&0.2863\\
        recall@4      &0.9762&0.8388&0.8405&0.9634&0.9726&0.9764&0.3064\\
        recall@16     &0.9789&0.8485&0.8506&0.9685&0.9759&0.9792&0.3257\\
        recall@64     &0.9800&0.8582&0.8589&0.9708&0.9773&0.9804&0.3489\\
        recall\_avg   &0.9757&0.8422&0.8430&0.9632&0.9721&0.9757&0.3168\\
        \hline
        \end{tabular}}
\end{table}

\begin{table}[tb]
    \centering
    \caption{details of contrastive learning training in OCR errors datasets}
    \label{contrastive learning training in OCR errors datasets}
        \resizebox{0.95\linewidth}{!}{
        \begin{tabular}{c|ccc|ccc|c}
        \hline
        \multicolumn{1}{c|}{batchsize}& \multicolumn{3}{c|}{64} & \multicolumn{3}{c|}{512} & \multirow{2}{*}{not train}   \\
        learning rate & 1e-5 & 3e-5 & 5e-5     & 1e-5 & 3e-5 & 5e-5      &                             \\
        \hline
        \multicolumn{1}{l|}{\emph{OCR\_1}} & & & & & & & \\
        recall@1      &0.9952&0.9976&0.9974&0.9971&0.9980&0.9981&0.9247\\
        recall@4      &0.9961&0.9982&0.9980&0.9978&0.9986&0.9987&0.9258\\
        recall@16     &0.9965&0.9984&0.9983&0.9981&0.9988&0.9990&0.9264\\
        recall@64     &0.9968&0.9986&0.9986&0.9983&0.9991&0.9992&0.9269\\
        recall\_avg   &0.9961&0.9982&0.9981&0.9978&0.9986&0.9987&0.9259\\
        \hline
        \multicolumn{1}{l|}{\emph{OCR\_2}} & & & & & & & \\
        recall@1      &0.9815&0.9899&0.9871&0.9901&0.9921&0.9919&0.8291\\
        recall@4      &0.9849&0.9921&0.9898&0.9924&0.9942&0.9939&0.8359\\
        recall@16     &0.9864&0.9930&0.9910&0.9936&0.9951&0.9948&0.8396\\
        recall@64     &0.9875&0.9937&0.9917&0.9943&0.9958&0.9955&0.8427\\
        recall\_avg   &0.9850&0.9922&0.9899&0.9926&0.9943&0.9940&0.8368\\
        \hline
        \multicolumn{1}{l|}{\emph{OCR\_3}} & & & & & & & \\
        recall@1      &0.9945&0.9970&0.9967&0.9967&0.9975&0.9975&0.9212\\
        recall@4      &0.9954&0.9976&0.9974&0.9973&0.9981&0.9982&0.9227\\
        recall@16     &0.9958&0.9979&0.9978&0.9976&0.9985&0.9985&0.9234\\
        recall@64     &0.9963&0.9983&0.9983&0.9981&0.9989&0.9989&0.9240\\
        recall\_avg   &0.9955&0.9977&0.9976&0.9974&0.9983&0.9983&0.9229\\
        \hline
        \multicolumn{1}{l|}{\emph{OCR\_4}} & & & & & & & \\
        recall@1      &0.9767&0.9866&0.9824&0.9869&0.9891&0.9889&0.8059\\
        recall@4      &0.9806&0.9893&0.9858&0.9897&0.9915&0.9914&0.8132\\
        recall@16     &0.9826&0.9905&0.9873&0.9913&0.9927&0.9927&0.8180\\
        recall@64     &0.9840&0.9915&0.9885&0.9923&0.9938&0.9937&0.8225\\
        recall\_avg   &0.9810&0.9895&0.9860&0.9901&0.9918&0.9917&0.8149\\
        \hline
        \multicolumn{1}{l|}{\emph{OCR\_mix}} & & & & & & & \\
        recall@1      &0.9869&0.9928&0.9907&0.9928&0.9942&0.9941&0.8694\\
        recall@4      &0.9892&0.9943&0.9926&0.9945&0.9956&0.9955&0.8736\\
        recall@16     &0.9903&0.9950&0.9934&0.9953&0.9964&0.9961&0.8761\\
        recall@64     &0.9912&0.9957&0.9941&0.9960&0.9970&0.9968&0.8782\\
        recall\_avg   &0.9894&0.9944&0.9927&0.9947&0.9958&0.9956&0.8743\\
        \hline
        \end{tabular}}
\end{table}

\section{the statistics of errors in the correction experiment}
\label{Appendix the statistics of errors in the correction experiment}
The statistics of errors in the correction experiment is as Table~\ref{the errors statistics of spelling mistakes on WNUT-17 dataset} and Table~\ref{the errors statistics of spelling mistakes on CoNLL-03 dataset}. The $ent$ is the number of entities , $ent\_true$ is the number of correct entities in which all words are correct, $ent\_rate$ is the ratio. The $token$, $token\_true$ and $token\_rate$ is similar for the words in the sentences.

\begin{table}[tb]
    \centering
    \caption{the errors statistics of spelling mistakes on WNUT-17 dataset}
    \label{the errors statistics of spelling mistakes on WNUT-17 dataset}
        \resizebox{1.0\linewidth}{!}{
        \begin{tabular}{c|ccc|ccc}
            \hline
            \rule{0pt}{12pt}
            noisy\_level & ent  & ent\_true & ent\_rate & token  & token\_true & token\_rate  \\
            \hline
            \multicolumn{1}{l|}{\emph{not correct}} & & & & & & \\
            0.1 & 3890 & 1743         & 44.81      & 101857 & 70194          & 68.91         \\
            0.2 & 3890 & 878          & 22.57      & 101857 & 51608          & 50.67         \\
            0.3 & 3890 & 510          & 13.11      & 101857 & 39014          & 38.30         \\
            \hline
            \multicolumn{1}{l|}{\emph{natas}} & & & & & & \\
            0.1 & 3890 & 878          & 22.57      & 101857 & 56533          & 55.50         \\
            0.2 & 3890 & 428          & 11.00      & 101857 & 44020          & 43.22         \\
            0.3 & 3890 & 243          & 6.25       & 101857 & 35338          & 34.69         \\
            \multicolumn{1}{l|}{\emph{hunspell}} & & & & & & \\
            0.1 & 3890 & 1695         & 43.57      & 101857 & 74196          & 72.84         \\
            0.2 & 3890 & 1181         & 30.36      & 101857 & 60803          & 59.69         \\
            0.3 & 3890 & 854          & 21.95      & 101857 & 49001          & 48.11         \\
            \hline
            \multicolumn{1}{l|}{\emph{natas+check}} & & & & & & \\
            0.1 & 3890 & 1759         & 45.22      & 101857 & 72845          & 71.52         \\
            0.2 & 3890 & 895          & 23.01      & 101857 & 55121          & 54.12         \\
            0.3 & 3890 & 519          & 13.34      & 101857 & 42741          & 41.96         \\
            \multicolumn{1}{l|}{\emph{hunspell+check}} & & & & & & \\
            0.1 & 3890 & 2399         & 61.67      & 101857 & 80038          & 78.58         \\
            0.2 & 3890 & 1535         & 39.46      & 101857 & 64160          & 62.99         \\
            0.3 & 3890 & 1053         & 27.07      & 101857 & 51169          & 50.24         \\
            \hline
        \end{tabular} }
\end{table}

\begin{table}[tb]
    \centering
    \caption{the errors statistics of spelling mistakes on CoNLL-03 dataset}
    \label{the errors statistics of spelling mistakes on CoNLL-03 dataset}
        \resizebox{1.0\linewidth}{!}{
        \begin{tabular}{c|ccc|ccc}
            \hline
            \rule{0pt}{12pt}
            noisy\_level & ent  & ent\_true & ent\_rate & token  & token\_true & token\_rate  \\
            \hline
            \multicolumn{1}{l|}{\emph{not correct}} & & & & & & \\
            0.1 & 35089 & 15145        & 43.16      & 301418 & 201883         & 66.98         \\
            0.2 & 35089 & 7118         & 20.29      & 301418 & 142756         & 47.36         \\
            0.3 & 35089 & 3448         & 9.83       & 301418 & 105795         & 35.10          \\
            \hline
            \multicolumn{1}{l|}{\emph{natas}} & & & & & & \\
            0.1 & 35089 & 7430         & 21.17      & 301418 & 173173         & 57.45         \\
            0.2 & 35089 & 3414         & 9.73       & 301418 & 129644         & 43.01         \\
            0.3 & 35089 & 1646         & 4.69       & 301418 & 100673         & 33.40         \\
            \multicolumn{1}{l|}{\emph{hunspell}} & & & & & & \\
            0.1 & 35089 & 17322        & 49.37      & 301418 & 230626         & 76.51         \\
            0.2 & 35089 & 12946        & 36.89      & 301418 & 186334         & 61.82         \\
            0.3 & 35089 & 9264         & 26.40       & 301418 & 148364        & 49.22         \\
            \hline
            \multicolumn{1}{l|}{\emph{natas+check}} & & & & & & \\
            0.1 & 35089 & 15149        & 43.17      & 301418 & 210017         & 69.68         \\
            0.2 & 35089 & 7122         & 20.30      & 301418 & 153086         & 50.79         \\
            0.3 & 35089 & 3450         & 9.83       & 301418 & 115691         & 38.38         \\
            \hline
            \multicolumn{1}{l|}{\emph{hunspell+check}} & & & & & & \\
            0.1 & 35089 & 22755        & 64.85      & 301418 & 242080         & 80.31         \\
            0.2 & 35089 & 15556        & 44.33      & 301418 & 193167         & 64.09         \\
            0.3 & 35089 & 10536        & 30.03      & 301418 & 152582         & 50.62         \\
            \hline
        \end{tabular} }
\end{table}

\printcredits

\bibliographystyle{cas-model2-names}

\bibliography{add_bib, base_bib}

\end{document}